\documentclass[sigconf]{acmart}

\copyrightyear{2026}
\acmYear{2026}
\setcopyright{cc}
\setcctype{by}
\acmConference[KDD 2026] {Proceedings of the 32nd ACM SIGKDD Conference on Knowledge Discovery and Data Mining V.2}{August 9--13, 2026}{Jeju Island, Republic of Korea.}
\acmBooktitle{Proceedings of the 32nd ACM SIGKDD Conference on Knowledge Discovery and Data Mining V.2 (KDD 2026), August 9--13, 2026, Jeju Island, Republic of Korea}
\acmISBN{979-8-4007-2259-2/2026/08}
\acmDOI{10.1145/3770855.3818079}

\usepackage{booktabs}
\usepackage{subcaption}
\usepackage{enumitem}
\usepackage{tikz}
\usetikzlibrary{shapes, arrows, positioning, calc}
\usepackage{algorithm}
\usepackage{algorithmic}

\AtBeginDocument{}

\title{SL-BiLEM: Structured Learnable Behavior-in-the-Loop Epidemic Modeling for Forecasting and Policy Evaluation}

\author{Haochun Wang}
\affiliation{%
  \department{Faculty of Computing}
  \institution{Harbin Institute of Technology}
  \city{Harbin}
  \country{China}
}
\email{hcwang@ir.hit.edu.cn}

\author{Sendong Zhao}
\authornote{Corresponding author.}
\affiliation{%
  \department{Faculty of Computing}
  \institution{Harbin Institute of Technology}
  \city{Harbin}
  \country{China}
}
\email{sdzhao@ir.hit.edu.cn}

\author{Jingbo Wang}
\affiliation{%
  \department{Faculty of Computing}
  \institution{Harbin Institute of Technology}
  \city{Harbin}
  \country{China}
}
\email{jbwang@ir.hit.edu.cn}

\author{Yanrui Du}
\affiliation{%
  \department{Faculty of Computing}
  \institution{Harbin Institute of Technology}
  \city{Harbin}
  \country{China}
}
\email{yrdu@ir.hit.edu.cn}

\author{Ting Liu}
\affiliation{%
  \department{Faculty of Computing}
  \institution{Harbin Institute of Technology}
  \city{Harbin}
  \country{China}
}
\email{tliu@ir.hit.edu.cn}

\author{Bing Qin}
\affiliation{%
  \department{Faculty of Computing}
  \institution{Harbin Institute of Technology}
  \city{Harbin}
  \country{China}
}
\email{qinb@ir.hit.edu.cn}

\renewcommand{\shortauthors}{Haochun Wang et al.}

\begin{document}

\begin{abstract}
Epidemic forecasting faces a fundamental challenge: human behavior dynamically responds to disease spread, creating feedback loops that induce distribution shifts at policy intervention points. This renders data-driven models unreliable under distribution shift. We propose \textbf{SL-BiLEM} (Structured Learnable Behavior-in-the-Loop Epidemic Model), leveraging physical constraints as regularization for robust extrapolation. The framework decomposes effective transmission as $\beta_{\text{eff}}(t,g) = \beta_0(g) \times m_{\text{policy}}(t) \times m_{\text{media}}(t) \times m_{\text{comp}}(t,g)$, where monotonicity, smoothness, and bounded-jump constraints on the learned compliance function maintain predictive validity under novel policy regimes. Beyond forecasting, SL-BiLEM enables counterfactual analysis for intervention decision support. We validate forecasting on three real-world datasets (cruise ship, school influenza, and school-district COVID-19 surveillance) and evaluate counterfactual recovery on synthetic benchmarks with known ground truth. SL-BiLEM demonstrates: (1) 76\% improvement over neural-mechanistic baselines, with only 53\% OOD degradation versus 1142\% for neural baselines under policy-induced shift; (2) 100\% bootstrap CI coverage across 27 synthetic counterfactual experiments; and (3) Treatment Effect Accuracy exceeding 0.85. These results establish SL-BiLEM as an interpretable tool for public health decision-makers seeking accurate prediction and principled intervention planning.

\end{abstract}




\maketitle

\section{Introduction}

Epidemic forecasting and policy evaluation are critical for public health decision-making, yet they represent fundamentally different analytical tasks. \emph{Forecasting} asks ``what will happen?''---predicting future case counts given current trends. \emph{Policy evaluation} asks ``what would happen if?''---estimating counterfactual outcomes under alternative intervention scenarios. The COVID-19 pandemic, which caused over 7 million deaths globally and trillions of dollars in economic losses~\cite{flaxman2020}, demonstrated the urgent need for both capabilities: accurate short-term forecasts for resource allocation, and credible counterfactual analysis for intervention timing and policy design. However, existing methods excel at one task while failing at the other, leaving decision-makers without a unified tool for evidence-based planning.

The fundamental challenge underlying both tasks is behavior-in-the-loop feedback: human behavior dynamically responds to disease spread, creating feedback loops that alter transmission dynamics~\cite{funk2010modelling,weitz2020pnas}. As case counts rise, individuals adapt through policy compliance, risk perception, and information exposure---wearing masks, reducing contacts, or avoiding crowded spaces. These behavioral changes reduce transmission, which in turn affects future case counts, completing the feedback loop. Critically, this feedback induces \emph{distribution shifts} precisely when reliable predictions matter most at policy intervention points. When a government announces a lockdown, the statistical relationships learned from pre-intervention data no longer hold, causing data-driven models to fail catastrophically. This is not merely a technical inconvenience---it represents a fundamental barrier to trustworthy epidemic modeling.

Current epidemic modeling paradigms each address only part of this challenge. \textbf{Compartmental models} (e.g., SEIR~\cite{kermack1927,anderson1992}) provide interpretable structure with natural counterfactual semantics---one can simulate ``what if $\beta$ were different?''---but they absorb behavioral dynamics into hand-crafted time-varying transmission rates $\beta(t)$, typically specified as step functions or threshold-based rules. These manual specifications cannot capture complex, data-driven behavioral responses and are difficult to validate or transfer across settings. \textbf{Data-driven models} including time-series forecasters (ARIMA, Prophet, neural networks) can extrapolate trends with high in-sample accuracy, but they lack explicit representations of interventions. Under policy-induced distribution shifts, their predictions degrade severely, and they cannot provide counterfactual estimates because their learned correlations conflate causal and spurious relationships~\cite{reich2022}. \textbf{LLM-based approaches}~\cite{williams2023llm} offer flexibility in incorporating textual information, but using LLMs as black-box forecasters raises fundamental concerns about reproducibility, bias, and auditability---critical requirements for public health applications where decisions affect millions of lives. No existing approach simultaneously achieves: (i) learnable behavioral dynamics from data, (ii) interpretable structure for counterfactual analysis, (iii) robustness under distribution shift, and (iv) reproducibility for scientific validation.

To address these limitations, we propose \textbf{SL-BiLEM} (\textbf{S}tructured \textbf{L}earnable \textbf{B}ehavior-\textbf{i}n-the-\textbf{L}oop \textbf{E}pidemic \textbf{M}odel), a unified framework that simultaneously enables multi-horizon forecasting and counterfactual policy evaluation. The key insight is that behavioral responses, while complex, obey certain physical regularities: compliance increases monotonically with perceived risk, changes smoothly over time, and exhibits bounded daily variations. By encoding these regularities as inductive biases, we constrain the hypothesis space to behaviorally plausible functions that generalize under distribution shift. Specifically, we decompose effective transmission into interpretable multiplicative factors:
\begin{equation*}
\beta_{\text{eff}}(t,g) = \beta_0(g) \times m_{\text{policy}}(t) \times m_{\text{media}}(t) \times m_{\text{comp}}(t,g)
\end{equation*}
where $\beta_0(g)$ is the baseline transmissibility for group $g$, $m_{\text{policy}}(t)$ captures direct policy effects derived from intervention signals, $m_{\text{media}}(t)$ captures information-driven behavioral changes, and $m_{\text{comp}}(t,g)$ is a learned compliance function that responds to epidemic risk under physical constraints (monotonicity, smoothness, bounded jumps). The term ``structured'' means each factor has clear epidemiological semantics and can be independently analyzed or ablated; ``learnable'' means $m_{\text{comp}}$ is calibrated from case data rather than hand-specified. To incorporate real-world policy information without sacrificing reproducibility, we restrict LLMs to upstream text-to-event extraction, ensuring that downstream mechanistic modeling and counterfactual assumptions remain transparent and auditable.

In summary, our contributions are as follows:
\begin{enumerate}
\item \textbf{Structured learnable compliance with physical constraints.} We propose a novel decomposition of effective transmission into interpretable multiplicative factors, where the compliance function is learned from case data under three physical constraints---monotonicity, smoothness, and bounded jumps. Unlike hand-crafted $\beta(t)$ specifications in traditional compartmental models, these constraints serve as inductive biases that enable robust extrapolation under distribution shift while maintaining epidemiological interpretability.

\item \textbf{Reproducible LLM integration for policy signal extraction.} We restrict LLMs to upstream text-to-event extraction, isolating LLM variability to a single cacheable preprocessing step while preserving mechanistic transparency for downstream counterfactual analysis. This architectural choice ensures that intervention assumptions remain explicit and auditable---a critical requirement for public health applications.

\item \textbf{Comprehensive empirical validation with novel evaluation metrics.} We validate SL-BiLEM on three real-world datasets spanning diverse epidemic scenarios (cruise ship, school influenza, and school-district COVID-19 surveillance), demonstrating 76\% improvement over neural-mechanistic baselines with only 53\% OOD degradation versus 1142\% for neural baselines. We introduce Treatment Effect Accuracy (TEA) to quantify counterfactual estimation quality, achieving TEA $\geq 0.85$ with 100\% bootstrap CI coverage across 27 synthetic experiments with known ground truth.
\end{enumerate}

\section{Related Work}

\subsection{Behavior-in-the-Loop Epidemic Modeling}

Real-world epidemics exhibit feedback loops between human behavior and disease spread~\cite{funk2010modelling,ward2023bayesian}. Traditional models ignoring behavior change can misestimate dynamics and intervention effects. Behavioral models range from data-driven approaches using mobility data~\cite{google2020mobility} to analytical feedback models with endogenous mechanisms~\cite{weitz2020pnas}. The latter often outperform pure data-driven methods in capturing long-term dynamics, as they model the feedback loop rather than merely fitting historical trends~\cite{comparative2025pmc}. Studies show that behavioral sensitivity varies with cumulative deaths, vaccination rates, and ``pandemic fatigue''~\cite{weitz2020pnas,llm2025arxiv}. Most existing models either rely on hand-crafted $\beta(t)$ functions or require real-time mobility data for forecasting. Our framework addresses this by learning compliance from case data under structural constraints.

\subsection{Multi-Group Heterogeneity and Network-Based Models}

Different population groups exhibit heterogeneous contact patterns and susceptibilities~\cite{mossong2008social}. Age-stratified models using contact matrices (e.g., POLYMOD) have been crucial in COVID-19 and influenza studies, revealing that young adults often serve as transmission ``sources'' while elderly populations are infection ``sinks''~\cite{ferguson2006nature,prem2017projecting,recurrent2025medrxiv}. Mixing assumptions have evolved from homogeneous random mixing to semi-random mixing (SeRaMix) that acknowledges contacts are drawn from finite recurrent pools~\cite{recurrent2025medrxiv}. For fine-grained interventions, agent-based models (ABMs) simulate disease spread through explicit contact networks~\cite{kerr2021covasim}. Network topology critically affects epidemic dynamics: small-world networks show that even a few long-range links drastically lower the percolation threshold~\cite{watts1998collective}, while scale-free networks with hub nodes have no epidemic threshold in the infinite limit~\cite{pastorsatorras2001epidemic}. Modern ABMs model contacts as multi-layer networks including households, workplaces, and transient community contacts~\cite{covsyn2025medrxiv}. Our framework bridges compartmental efficiency with network-aware heterogeneity through structured mixing matrices and group-specific dynamics.

\subsection{Policy Intervention and Counterfactual Evaluation}

Epidemic outcomes are shaped by exogenous interventions and information feedback~\cite{adaptive2023mdm}. Adaptive policies use surveillance triggers with trade-offs: low thresholds squash peaks but require longer restrictions; high thresholds risk larger outbreaks~\cite{adaptive2023mdm,twogroup2025aims}. Coupled information-epidemic models analyze how official news versus rumors impact compliance, though excessive information may cause fatigue~\cite{funk2010modelling,coupled2024sciencedirect}. Large language models offer new possibilities for integrating unstructured data---extracting policy events from legal text, news, and WHO reports~\cite{williams2023llm,eventextraction2025acl}. However, using LLMs as direct predictors raises reproducibility concerns~\cite{llm2025arxiv}. Our framework restricts LLMs to upstream extraction with caching and optional distillation, ensuring the core model remains transparent. Counterfactual analysis faces the fundamental problem that we cannot observe both factual and counterfactual outcomes simultaneously~\cite{causalchallenges2025arxiv}. Physics-informed neural networks embed epidemiological equations into loss functions, ensuring predictions respect biological constraints~\cite{pisid2025pmc}. Synthetic datasets with known causal ground truth enable benchmarking~\cite{covsyn2025medrxiv}. However, existing work lacks standardized metrics for evaluating counterfactual estimation quality. 

\section{Problem Formulation}

We formalize the epidemic modeling problem with explicit consideration of behavioral feedback and distribution shift challenges.

\subsection{Notation and Data}

\paragraph{Observed data.}
Let $\{y_t\}_{t=1}^T$ denote the daily reported case counts for a population of size $N$. When population heterogeneity is available, we observe group-indexed counts $\{y_{t,g}\}_{g \in \mathcal{G}}$ where $\mathcal{G}$ is the set of groups.

\paragraph{Intervention and information signals.}
We consider two types of signals that influence transmission dynamics:
\begin{itemize}
    \item Policy signal $\{s_t\}_{t=1}^T$: A strictness measure $s_t \in [0,1]$ representing the intensity of non-pharmaceutical interventions (NPIs) at time $t$, where $s_t = 0$ indicates no restrictions and $s_t = 1$ indicates maximum restrictions (e.g., full lockdown).
    \item Media/information signal $\{m_t\}_{t=1}^T$: A measure $m_t \in [0,1]$ capturing the intensity of risk-related information exposure (e.g., news coverage, public health announcements).
\end{itemize}
These signals may be directly available as structured data or extracted from unstructured text via upstream processing.

\paragraph{Mixing structure (multi-group extension).}
When modeling heterogeneous populations, we specify a contact matrix $C \in \mathbb{R}^{|\mathcal{G}| \times |\mathcal{G}|}$, where $C_{gh}$ represents the relative contact frequency between groups $g$ and $h$. This enables analysis of age-stratified or spatially-structured epidemics with group-specific transmission dynamics.

\subsection{Core Challenge: Behavior-Induced Distribution Shift}

The central challenge in epidemic modeling is that human behavior responds to disease dynamics, creating a feedback loop:
\begin{equation}
\begin{split}
\text{Cases } y_t &\xrightarrow{\text{risk perception}} \text{Behavior } c_t \\
&\xrightarrow{\text{transmission reduction}} \text{Future cases } y_{t+k}
\end{split}
\label{eq:feedback_loop}
\end{equation}
This feedback induces distribution shift at precisely the moments when accurate predictions are most critical---policy intervention points. When a government announces new restrictions, the statistical relationships learned from pre-intervention data no longer hold, causing purely data-driven models to fail. Any effective modeling framework must explicitly account for this behavioral feedback mechanism.

\subsection{Task Definitions}
\label{counter-def}
We address two complementary tasks that together support public health decision-making:

\paragraph{Task 1: Multi-horizon forecasting.}
Given observations and signals up to time $\tau$, i.e., $\mathcal{D}_\tau = \{(y_t, s_t, m_t)\}_{t=1}^\tau$, predict case counts for the next $H$ days:
\begin{equation}
\text{Forecast: } \mathcal{D}_\tau \mapsto \{\hat{y}_{\tau+1}, \hat{y}_{\tau+2}, \ldots, \hat{y}_{\tau+H}\}.
\end{equation}
We evaluate forecasting performance in a rolling-origin setting: for each cut-off time $\tau \in \{\tau_1, \tau_2, \ldots\}$, we generate forecasts at multiple horizons $H \in \{7, 14, 28\}$ days and compare against held-out observations.

\paragraph{Task 2: Counterfactual policy evaluation.}
Given a fitted model on observed history $\mathcal{D}_T$ and a counterfactual intervention scenario $\{s_t^{\text{cf}}\}$ (e.g., earlier lockdown, different strictness level), estimate what would have happened under the alternative policy:
\begin{equation}
\text{Counterfactual: } (\theta^*, \{s_t^{\text{cf}}\}) \mapsto \{\hat{y}_t^{\text{cf}}\}_{t=1}^T,
\end{equation}
where $\theta^*$ denotes the calibrated model parameters. Key counterfactual quantities include:
\begin{itemize}
    \item Cumulative cases averted: $\sum_t (y_t^{\text{obs}} - \hat{y}_t^{\text{cf}})$
    \item Peak reduction: $(\max_t y_t^{\text{obs}} - \max_t \hat{y}_t^{\text{cf}}) / \max_t y_t^{\text{obs}}$
    \item Delay to peak: $\arg\max_t \hat{y}_t^{\text{cf}} - \arg\max_t y_t^{\text{obs}}$
\end{itemize}

\section{Method}


\subsection{Overview}
\label{sec:design_philosophy}

The design of SL-BiLEM is guided by two core principles:

\paragraph{Principle 1: Structured decomposition for interpretability.}
Rather than learning a monolithic time-varying transmission rate $\beta(t)$, we decompose effective transmission into semantically meaningful factors---policy effects, media influence, and behavioral compliance---enabling independent analysis, transparent counterfactual reasoning, and component transfer across settings.

\paragraph{Principle 2: Physical constraints as regularization.}
We encode behavioral regularities as hard constraints: compliance increases monotonically with perceived risk, changes smoothly over time, and exhibits bounded daily variations. These constraints dramatically reduce the hypothesis space, preventing overfitting to spurious correlations that break under distribution shift.

\begin{figure*}[t]
\centering
\Description{SL-BiLEM framework architecture.}
\includegraphics[width=0.8\textwidth]{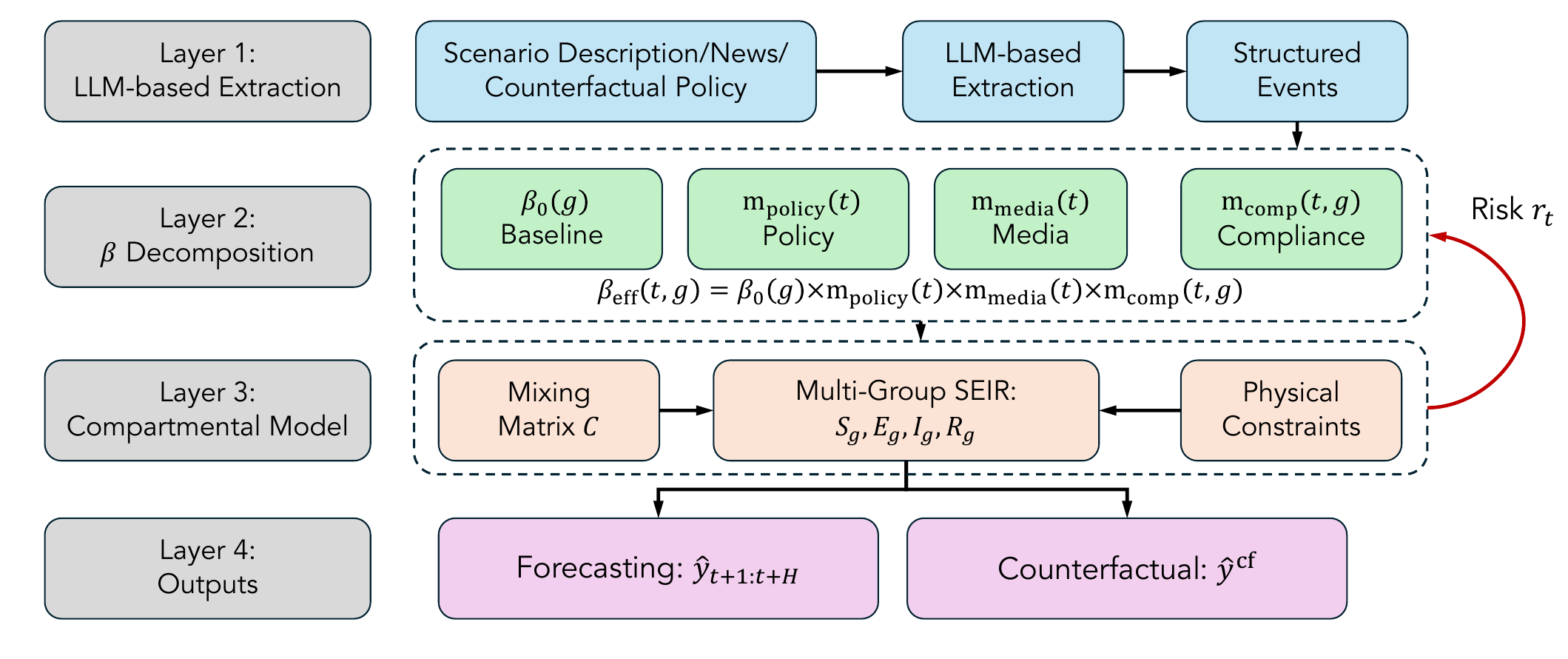}
\caption{SL-BiLEM framework architecture. The core SEIR dynamics are modulated by a structured transmission decomposition that includes a learnable compliance component. Physical constraints are enforced on the compliance function to ensure behavioral plausibility and robust extrapolation under distribution shift.}
\label{fig:framework}
\end{figure*}

\subsection{Mechanistic Foundation: SEIR Dynamics}
\label{sec:seir_dynamics}

We build on the classical SEIR compartmental model as our dynamical backbone, chosen for its interpretability and well-understood epidemiological semantics. For a single homogeneous population, the dynamics evolve susceptible $S(t)$, exposed $E(t)$, infectious $I(t)$, and removed $R(t)$ compartments:
\begin{align}
\frac{dS}{dt} &= -\beta_{\text{eff}}(t)\frac{S I}{N}, &
\frac{dE}{dt} &= \beta_{\text{eff}}(t)\frac{S I}{N} - \sigma E, \label{eq:seir1}\\
\frac{dI}{dt} &= \sigma E - \gamma I, &
\frac{dR}{dt} &= \gamma I, \label{eq:seir2}
\end{align}
where $N$ is population size, $\beta_{\text{eff}}(t)$ is the time-varying effective transmission rate (detailed in Section~\ref{sec:decomposition}), $\sigma$ is the incubation transition rate (inverse of mean latent period), and $\gamma$ is the recovery/removal rate (inverse of mean infectious period). We discretize these dynamics on a daily grid using a fourth-order Runge-Kutta integrator and treat daily incidence as $\hat{\imath}_t \approx \sigma E(t)$.
The key departure from standard SEIR is that $\beta_{\text{eff}}(t)$ is not a hand-crafted function but emerges from our structured decomposition with learned compliance, as detailed next.

\subsection{Structured Transmission Decomposition}
\label{sec:decomposition}

SL-BiLEM represents effective transmission as an interpretable multiplicative decomposition:
\begin{equation}
\beta_{\text{eff}}(t, g) = \beta_0(g) \cdot m_{\text{policy}}(t)\cdot m_{\text{media}}(t)\cdot m_{\text{comp}}(t, g),
\label{eq:beta_decomposition}
\end{equation}
where $\beta_0(g)$ is the baseline transmissibility for group $g$, and each multiplier $m \in [0,1]$ captures a distinct mechanism by which transmission is reduced. This decomposition is ``structured'' in the sense that each factor has clear epidemiological semantics. The multiplicative form reflects the epidemiological intuition that these factors act independently on contact rates.

Each multiplier takes the form $m = 1 - \rho \cdot x$, where $x \in [0,1]$ is a normalized input signal (policy strictness, media intensity, or compliance level) and $\rho \in [0,1]$ is a maximum reduction parameter that bounds the effect magnitude. We use subscripted $\rho$ notation consistently: $\rho_{\text{policy}}$, $\rho_{\text{media}}$, and $\rho_{\text{comp}}(g)$ for the three components respectively. This parameterization ensures: (i) $m = 1$ when $x = 0$ (no effect), (ii) $m \geq 1 - \rho$ even at maximum input (bounded effect), and (iii) interpretable effect sizes. The $\rho$ parameters encode domain knowledge about realistic effect ceilings---for instance, even perfect policy compliance cannot eliminate all transmission due to essential contacts. We now describe each component in detail.

\paragraph{Policy multiplier $m_{\text{policy}}(t)$.}
This multiplier captures the direct effect of non-pharmaceutical interventions (NPIs) such as lockdowns, school closures, or gathering restrictions. Following the multiplicative NPI effectiveness framework established in large-scale intervention studies~\cite{flaxman2020,brauner2021}, we model policy effects as proportional reductions in transmission. Given policy events with associated strictness levels $s_t\in[0,1]$ and a maximum reduction parameter $\rho_{\text{policy}}\in[0,1]$:
\begin{equation}
m_{\text{policy}}(t) = 1 - \rho_{\text{policy}}\, s_t.
\label{eq:policy_multiplier}
\end{equation}
Here $s_t = 0$ indicates no restrictions (full transmission), $s_t = 1$ indicates maximum restrictions, and $\rho_{\text{policy}}$ captures the effectiveness ceiling of policy interventions. This linear form is consistent with empirical findings that NPI effects are approximately additive in log-transmission space~\cite{brauner2021}. Policy events can be manually specified from official announcements or extracted from text using the LLM interface (Section~\ref{sec:llm_extraction}).

\paragraph{Media multiplier $m_{\text{media}}(t)$.}
This multiplier captures information-driven behavioral changes from media exposure that occur independently of mandated policies---for example, voluntary mask-wearing after news coverage of rising cases. Following coupled information-epidemic models~\cite{funk2010modelling,coupled2024sciencedirect}, we model media effects with exponential decay to capture two well-documented phenomena: (i) information persistence, where behavioral changes outlast the triggering event, and (ii) information fatigue, where the impact of news diminishes over time:
\begin{align}
a_t &= \min\left(1, \sum_{\substack{e \in \mathcal{E}_{\text{media}} \\ t_e \leq t}} I_e \cdot \exp\left(-\frac{t - t_e}{\tau_{\text{decay}}}\right)\right), \label{eq:media_signal}\\
m_{\text{media}}(t) &= 1 - \rho_{\text{media}} a_t.
\label{eq:media_multiplier}
\end{align}
where $\mathcal{E}_{\text{media}}$ is the set of media events, $I_e \in [0,1]$ is the intensity of event $e$, $t_e$ is the event time, $\tau_{\text{decay}}$ is the decay constant (default 14 days based on typical news cycle duration~\cite{coupled2024sciencedirect}), and $\rho_{\text{media}}$ is the maximum media-driven reduction. The clipping step keeps the accumulated media signal in $[0,1]$ and prevents the multiplier from becoming negative.

\paragraph{Compliance multiplier $m_{\text{comp}}(t, g)$.}
Unlike policy and media effects, which are driven by external events, compliance represents the behavioral response to perceived epidemic risk. This formulation is grounded in awareness-driven behavior change models~\cite{weitz2020pnas}, which demonstrate that epidemic dynamics can shift from peaks to plateaus when populations adaptively respond to perceived risk. The feedback loop---where rising cases increase compliance, which reduces transmission, which affects future cases---is the core mechanism that distinguishes behavior-in-the-loop models from static approaches~\cite{funk2010modelling,ward2023bayesian}. We model this as:
\begin{equation}
c_t = f_{\theta}\!\left(r_t,\, s_t\right), \qquad m_{\text{comp}}(t, g) = 1 - \rho_{\text{comp}}(g)\, c_{t,g},
\label{eq:compliance}
\end{equation}
where all quantities are normalized to $[0,1]$: $r_t$ is the risk signal computed from recent incidence (Appendix~\ref{sec:implementation}), $c_t$ is the compliance level, and $\rho_{\text{comp}}(g)$ controls the maximum behavioral reduction for group $g$.
The function $f_{\theta}$ is the learnable compliance module optimized from case data under physical constraints, detailed next.

\subsection{Compliance with Physical Constraints}
\label{sec:learnable_compliance}

Learning the compliance function $f_{\theta}$ from case data is challenging because: (i) compliance is not directly observed---we only see its downstream effect on case counts; (ii) the feedback loop creates confounding between compliance and transmission; (iii) limited data during early outbreaks makes overfitting a severe risk. We impose physical constraints that encode behavioral regularities, reducing the hypothesis space while maintaining expressiveness.

\paragraph{Constraint 1: Monotonicity.}
Compliance should increase, or at least not decrease, with perceived risk, since people do not become less cautious as danger increases. Formally, $f_{\theta}$ must be monotonically non-decreasing in the risk signal. This constraint eliminates spurious solutions where the model explains transmission reduction through decreasing compliance, which would be behaviorally implausible.

\paragraph{Constraint 2: Smoothness.}
Behavioral changes exhibit inertia---habits, social norms, and cognitive limitations prevent instantaneous adaptation. We penalize rapid day-to-day changes:
\begin{equation}
\mathcal{R}_{\text{smooth}} = \lambda_s \sum_t \left(m_{\text{comp}}(t+1) - m_{\text{comp}}(t)\right)^2.
\label{eq:smoothness}
\end{equation}
This regularization prevents the model from fitting noise in daily case counts through high-frequency compliance oscillations.

\paragraph{Constraint 3: Bounded jumps.}
Even under dramatic events (e.g., celebrity death from COVID, major policy announcement), behavioral change has cognitive and social limits. We impose:
\begin{equation}
\left|m_{\text{comp}}(t+1) - m_{\text{comp}}(t)\right| \leq \delta_{\max},
\label{eq:bounded_jump}
\end{equation}
where $\delta_{\max}$ represents the maximum plausible daily change in compliance-driven transmission reduction.

\begin{figure}[t]
  \centering
  \Description{Two-panel figure showing compliance dynamics over time and the learned monotonic compliance function.}
  \includegraphics[width=\linewidth]{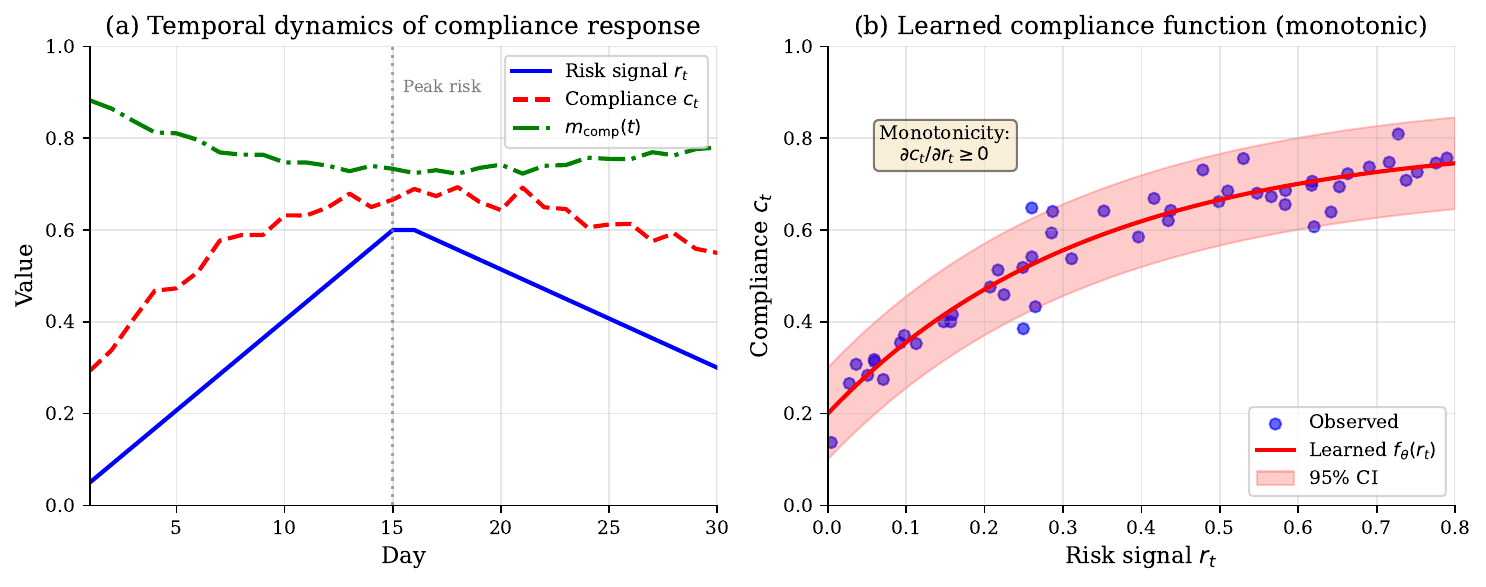}
  \caption{Schematic illustration of compliance behavior under physical constraints. (a) Temporal co-evolution of risk signal, compliance, and transmission multiplier. (b) The learned monotonic compliance function $f_\theta(r_t)$.}
  \label{fig:compliance_curve}
\end{figure}

\paragraph{Why these constraints enable robust extrapolation.}
The key insight is that these constraints encode domain knowledge across distribution shift. When a new policy is announced, the statistical correlations in historical data may break, but the behavioral regularities persist: people will still respond monotonically to risk, change gradually, and have bounded adaptation rates. By constraining the learned function to this behaviorally plausible subspace, we achieve models that extrapolate reliably to novel regimes.
Figure~\ref{fig:compliance_curve} provides a schematic illustration of a learned compliance function satisfying these constraints. The monotonic, smooth curve captures the intuitive relationship between risk and compliance while avoiding overfitting to noise.

The compliance function $f_\theta$ is implemented as a monotonic neural network with non-negative weight constraints to guarantee the monotonicity property. Architecture details are provided in Appendix~\ref{sec:appendix_architecture}.

\paragraph{Constrained Optimization}
\label{sec:optimization}
We enforce the physical constraints via projected gradient descent. The optimization objective combines a negative binomial likelihood with constraint penalties:
\begin{equation}
\min_\theta \mathcal{L}(\theta) + \lambda_s \mathcal{R}_{\text{smooth}}(\theta) + \lambda_m \mathcal{R}_{\text{mono}}(\theta),
\label{eq:objective}
\end{equation}
where:
\begin{equation}
\mathcal{L}(\theta) = -\sum_t \log P_{\text{NB}}(y_t \mid \mu_t(\theta), r),
\label{eq:likelihood}
\end{equation}
$y_t$ is the observed case count, $\mu_t(\theta)$ is the model-predicted expected count, $r$ is the overdispersion parameter, and $\mathcal{R}_{\text{mono}}$ penalizes violations of monotonicity.
At each optimization step, we project onto the feasible set via sequential operations: (i) non-negative weight clipping for monotonicity, (ii) isotonic regression for smoothness, and (iii) sequential clipping for bounded jumps. The complete training procedure is provided in Algorithm~\ref{alg:training} (Appendix~\ref{sec:appendix_algorithms}).

\subsection{Multi-Group Extension}
\label{sec:multigroup}

Real epidemics exhibit heterogeneity across demographic groups (age, occupation, location) with distinct contact patterns and behavioral responses. SL-BiLEM extends naturally to heterogeneous populations.

\paragraph{Group-specific dynamics.}
Each group $g \in \mathcal{G}$ maintains its own SEIR compartments with group-specific parameters:
\begin{align}
\frac{dS_g}{dt} &= -S_g \sum_{h \in \mathcal{G}} C_{gh} \beta_{\text{eff}}(t, h) \frac{I_h}{N_h}, \label{eq:multigroup1}\\
\frac{dE_g}{dt} &= S_g \sum_{h \in \mathcal{G}} C_{gh} \beta_{\text{eff}}(t, h) \frac{I_h}{N_h} - \sigma E_g, \label{eq:multigroup2}\\
\frac{dI_g}{dt} &= \sigma E_g - \gamma I_g, \quad \frac{dR_g}{dt} = \gamma I_g, \label{eq:multigroup3}
\end{align}
where $C_{gh}$ is the mixing matrix element representing relative contact frequency between groups $g$ and $h$. Each group has its own baseline transmissibility $\beta_0(g)$, maximum compliance reduction $\rho_{\text{comp}}(g)$, and compliance response $c_{t,g}$.



\subsection{Counterfactual Policy Evaluation}
\label{sec:counterfactual}

Counterfactual analysis estimates what would have happened under alternative policy scenarios. This is enabled by the structured decomposition, which allows modifying specific components while holding others fixed.
A counterfactual scenario specifies alternative policy events $\{s'_t\}$ while holding epidemiological parameters $(\beta_0, \sigma, \gamma)$ and compliance structure $f_\theta$ fixed at their calibrated values. The counterfactual trajectory is obtained by re-simulating with modified policy multipliers:
\begin{equation}
m'_{\text{policy}}(t) = 1 - \rho_{\text{policy}} \cdot s'_t.
\label{eq:counterfactual_policy}
\end{equation}

We compute the three policy-relevant metrics defined in Section~\ref{counter-def}. To quantify uncertainty, we use block bootstrap: resample and refit the model $B$ times, compute counterfactual trajectories for each bootstrap sample, and report 95\% confidence intervals from the empirical distribution. The complete counterfactual evaluation procedure is provided in Algorithm~\ref{alg:counterfactual} (Appendix~\ref{sec:appendix_algorithms}). 

\subsection{LLM-Based Event Extraction}
\label{sec:llm_extraction}

Policy and media events can be extracted from unstructured text using LLMs. Each extracted event contains: event type (policy/media), date, strictness/intensity $\in[0,1]$, affected groups, and description. To ensure reproducibility, we restrict LLMs to this upstream extraction step and provide caching mechanisms for deterministic replay. Implementation details are in Appendix~\ref{sec:appendix_llm}.

\section{Experiments}


\subsection{Datasets}

We use three real-world datasets representing diverse epidemic scenarios:

\begin{itemize}[leftmargin=*,nosep]
\item \textbf{Diamond Princess}~\cite{mizumoto2020diamond}: 14-day COVID-19 outbreak (Feb 5--19, 2020) on a cruise ship with 3,711 passengers and crew. The known quarantine timing provides a natural policy intervention point, making this an ideal testbed for evaluating policy-aware models.

\item \textbf{British Boarding School}~\cite{cdc1978influenza}: Classic 14-day H1N1 outbreak (763 students, Jan 22--Feb 4, 1978). No explicit policy intervention was implemented, but natural behavioral response to rising cases creates an implicit feedback loop.

\item \textbf{Illinois Schools}~\cite{smith2024shield}: Illinois K--8 school-district COVID-19 surveillance data with 8,150 total population (6,997 students and 1,153 employees), featuring repeated testing and intervention records including mask mandates, hybrid learning transitions, and quarantine protocols.
\end{itemize}

Dataset details are provided in Appendix~\ref{sec:appendix_dataset_details}.

\subsection{Baselines and Evaluation Protocol}

We compare against three categories of baselines representing the spectrum of epidemic modeling approaches:

\paragraph{Mechanistic models.}
SEIR (constant $\beta$, no behavioral response), SEIR+Policy (step-function reduction after known intervention time $t_p$), and SEIR+Threshold (binary response when cases exceed learned threshold $\theta$). These represent progressively sophisticated approaches to behavioral modeling: SEIR+Policy captures discrete regime changes from external interventions, while SEIR+Threshold captures reactive behavior triggered by epidemic severity. Both support counterfactual analysis but assume abrupt, binary responses rather than the continuous adaptation observed in real populations.

\paragraph{Data-driven models.}
Prophet~\cite{taylor2018prophet} (additive time series decomposition) and TCN~\cite{bai2018tcn} (temporal convolutional network). These methods achieve strong in-distribution forecasting through pattern matching but cannot perform counterfactual analysis.

\paragraph{Hybrid neural-mechanistic models.}
Neural ODE~\cite{chen2018neural} (continuous-time dynamics) and EARTH~\cite{wan2025earth} (neural ODEs with implicit SIR-like state transitions and graph-based regional transmission). While incorporating mechanistic structure, these methods lack explicit behavioral feedback and cannot incorporate policy signals for counterfactual analysis.


\subsection{Forecasting Performance}

\subsubsection{Main Results}

Table~\ref{tab:forecast_all} demonstrates forecasting performance across all datasets and methods.

\begin{table}[htb]
  \caption{Forecasting performance (Test RMSE $\pm$ std) with 5-fold rolling-origin evaluation. Best results in \textbf{bold}, second best \underline{underlined}.}
  \label{tab:forecast_all}
  \centering
  \small
  \resizebox{\columnwidth}{!}{%
  \begin{tabular}{llrrr}
    \toprule
    Method & Category & Diamond P. & British S. & Illinois \\
    \midrule
    SEIR & Mechanistic & 102.23{\scriptsize$\pm$12.4} & 59.65{\scriptsize$\pm$8.2} & 22.46{\scriptsize$\pm$3.1} \\
    SEIR+Policy & Mechanistic & 43.31{\scriptsize$\pm$6.8} & 18.99{\scriptsize$\pm$3.4} & 9.95{\scriptsize$\pm$1.8} \\
    SEIR+Threshold & Mechanistic & 45.09{\scriptsize$\pm$7.1} & 18.79{\scriptsize$\pm$3.2} & \textbf{6.93}{\scriptsize$\pm$1.2} \\
    Prophet & Data-driven & 62.27{\scriptsize$\pm$9.3} & 97.10{\scriptsize$\pm$14.6} & 10.76{\scriptsize$\pm$2.1} \\
    TCN & Data-driven & \underline{38.32}{\scriptsize$\pm$5.2} & \underline{9.78}{\scriptsize$\pm$1.8} & 7.32{\scriptsize$\pm$1.4} \\
    Neural ODE & Hybrid & 163.31{\scriptsize$\pm$21.7} & 56.75{\scriptsize$\pm$9.8} & 108.31{\scriptsize$\pm$15.2} \\
    EARTH & Hybrid & 52.16{\scriptsize$\pm$7.8} & 15.42{\scriptsize$\pm$2.9} & 8.85{\scriptsize$\pm$1.6} \\
    \textbf{SL-BiLEM} & Ours & \textbf{34.19}{\scriptsize$\pm$5.1} & \textbf{3.69}{\scriptsize$\pm$0.8} & \underline{7.12}{\scriptsize$\pm$1.1} \\
    \bottomrule
  \end{tabular}%
  }
\end{table}

SL-BiLEM achieves the best RMSE on Diamond Princess and British Boarding School, while achieving competitive second-best performance on Illinois Schools (7.12 vs.\ 6.93 for SEIR+Threshold). Notably, SL-BiLEM outperforms all data-driven methods (TCN, Prophet) and hybrid neural-mechanistic approaches (EARTH, Neural ODE) across all datasets.

The performance gap is most striking on British Boarding School, where SL-BiLEM achieves a 76\% improvement over EARTH (3.69 vs.\ 15.42). This dataset lacks explicit policy interventions, making behavioral response to rising cases the dominant dynamic. SL-BiLEM's explicit compliance mechanism captures this feedback loop, whereas EARTH has three limitations in this setting: (1) it requires multi-regional data with geographic adjacency structure, making it inapplicable to single-location outbreaks without artificial graph construction; (2) its implicit SIR-like state transitions cannot incorporate external policy signals; and (3) it cannot perform counterfactual analysis because its learned dynamics are correlational rather than causal.

\subsubsection{Out-of-Distribution Robustness}

A critical question for real-world deployment is how models perform when test conditions differ from training. We design a controlled OOD experiment where training data contains no extreme policy interventions ($s_t \leq 0.3$, advisory-level only), but test data includes a sudden strict lockdown ($s_t = 0.9$) starting on day 21.

\begin{figure}[t]
  \centering
  \Description{Bar chart comparing in-distribution and out-of-distribution RMSE for different methods.}
  \includegraphics[width=\linewidth]{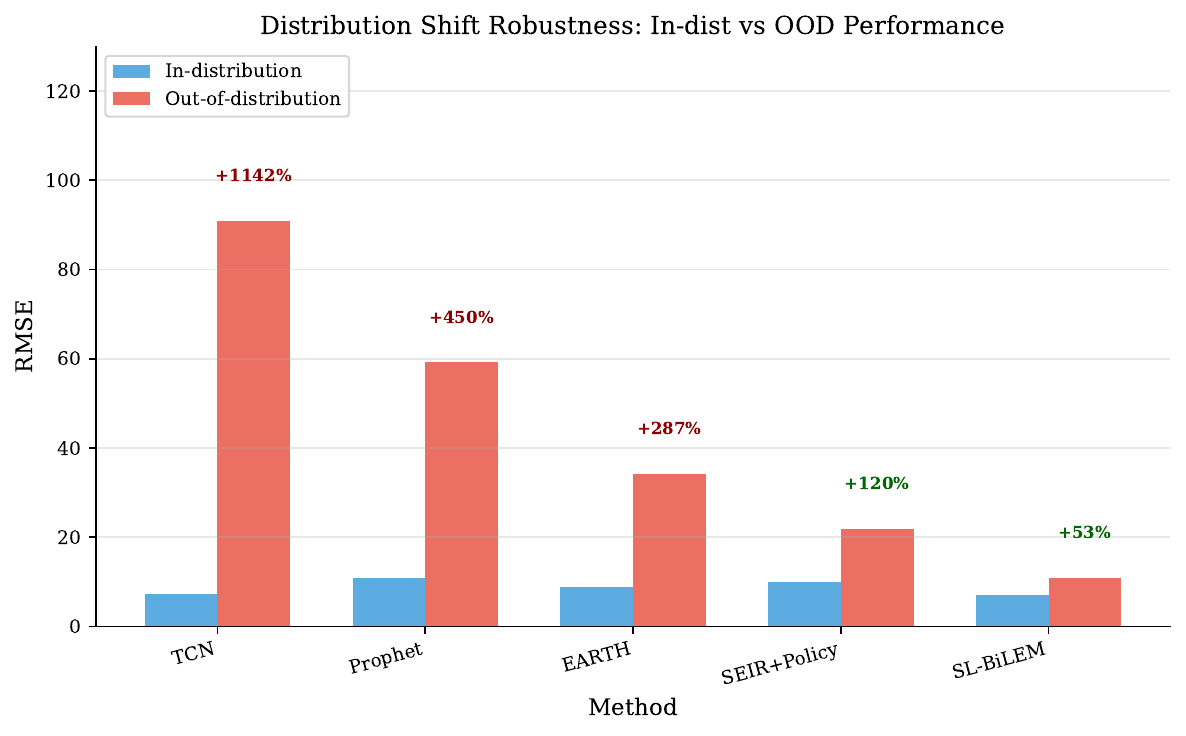}
  \caption{Distribution shift robustness: TCN achieves the best in-distribution RMSE but fails under distribution shift (+1142\%), while SL-BiLEM degrades gracefully (+53\%).}
  \label{fig:ood_degradation}
\end{figure}

Figure~\ref{fig:ood_degradation} visualizes the robustness comparison. TCN achieves the best in-distribution RMSE but degrades by 1142\% under distribution shift, while SL-BiLEM degrades by only 53\%. This demonstrates three key findings: (1) in-distribution accuracy is a poor proxy for real-world utility---TCN's advantage disappears when policies change; (2) mechanistic structure provides inductive bias for unseen regimes: SEIR+Policy (+120\%) and SL-BiLEM (+53\%) degrade less than data-driven methods; (3) explicit behavioral constraints are necessary: EARTH still degrades by 287\% despite physics-inspired structure, while SL-BiLEM's compliance mechanism enables robust extrapolation. Detailed numerical results are in Table~\ref{tab:ood_results} (Appendix).

\subsubsection{Multi-Horizon Performance}

Across 7/14/28-day horizons on Illinois Schools, SL-BiLEM maintains stable performance while TCN's advantage at short horizons disappears at 28-day forecasts due to policy-induced distribution shift (Figure~\ref{fig:multi_horizon}). Extended results including temporal transfer experiments are in Appendix~\ref{sec:appendix_extended_forecast}.

\begin{figure}[htb]
  \centering
  \Description{Bar chart comparing RMSE across forecast horizons.}
  \includegraphics[width=\linewidth]{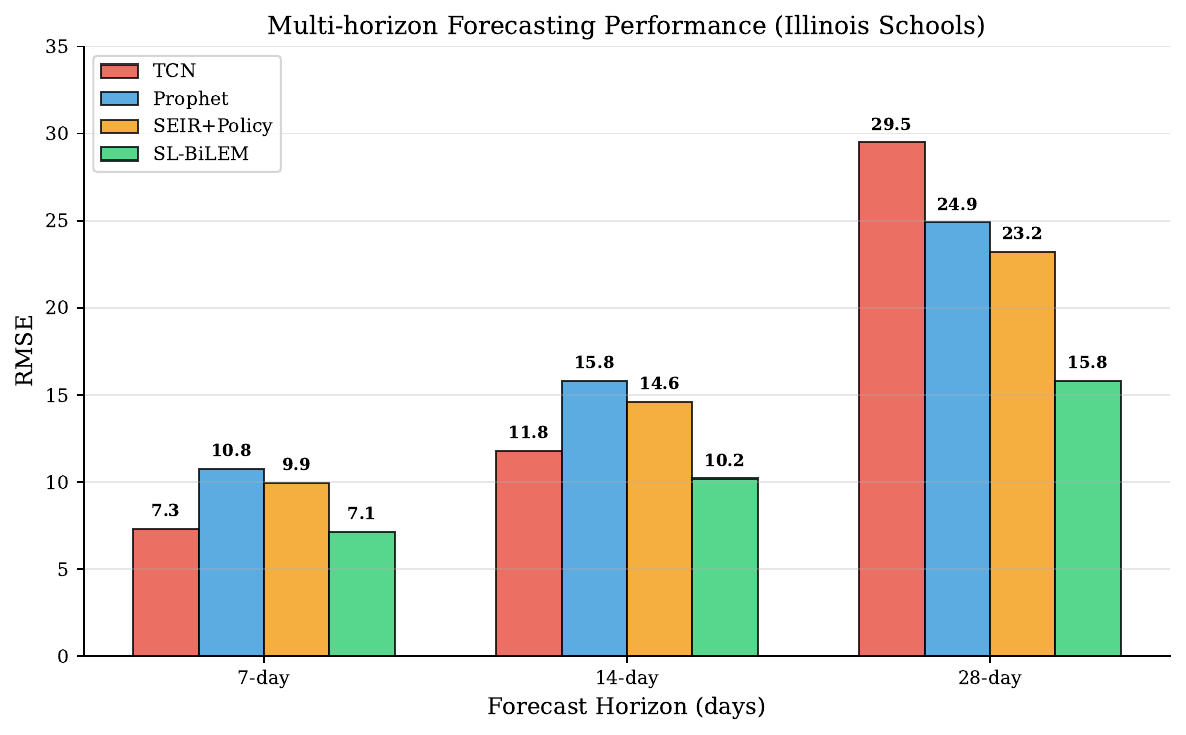}
  \caption{Multi-horizon forecasting on Illinois Schools. TCN excels at short horizons but degrades at 28-day; SL-BiLEM maintains stable performance across all horizons.}
  \label{fig:multi_horizon}
\end{figure}

\subsection{Counterfactual Analysis}

A key advantage of SL-BiLEM over data-driven methods is its ability to answer ``what-if'' questions about policy interventions. Table~\ref{tab:counterfactual_scenarios} summarizes counterfactual effects across synthetic policy-timing scenarios.

\begin{table}[htb]
  \caption{Counterfactual analysis: percentage change in cumulative cases under different policy timing relative to observed trajectory.}
  \label{tab:counterfactual_scenarios}
  \centering
  \small
  \begin{tabular}{lrrrr}
    \toprule
    Scenario & -7 days & -3 days & +3 days & No Policy \\
    \midrule
    Diamond Princess & -34.9\% & -17.1\% & +20.7\% & +184.3\% \\
    School Outbreak & N/A & -23.5\% & +24.8\% & +101.9\% \\
    Community Spread & -23.7\% & -10.9\% & +12.6\% & +282.3\% \\
    \bottomrule
  \end{tabular}
\end{table}

\paragraph{Policy timing is critical.}
Early intervention yields substantial benefits: implementing policy 7 days earlier reduces cumulative cases by 23--35\%. Conversely, delays are costly: 3-day delays increase cases by 12--25\%. Complete policy removal increases cases by 100--280\%, quantifying the substantial impact of behavioral interventions.

\paragraph{Synthetic ground-truth validation.}
To rigorously validate counterfactual estimation, we construct synthetic datasets using an agent-based model (ABM) with known causal effects---a structurally different data-generating process from SL-BiLEM. We introduce Treatment Effect Accuracy (TEA):
\begin{equation}
\text{TEA} = \max\left(0, 1 - \frac{|\hat{\tau} - \tau|}{|\tau|}\right)
\label{eq:tea}
\end{equation}
where $\tau$ is the true treatment effect and $\hat{\tau}$ is the model's estimate. On synthetic benchmarks, SL-BiLEM achieves TEA $\geq 0.85$ in 27/27 experiments with 100\% bootstrap CI coverage. Importantly, these CIs are informative rather than trivially wide: with a weekly block length ($b=7$), the average CI width is 23.4\% of the true effect magnitude (range: 15.2\%--34.7\%), indicating well-calibrated uncertainty quantification. We use moving block bootstrap to account for temporal dependence in epidemic time series~\cite{hall1995block}; the counterfactual procedure is detailed in Appendix~\ref{sec:appendix_algorithms}. Sensitivity analysis across block lengths $b \in \{3, 5, 7, 10\}$ shows stable coverage (96\%--100\%) with expected width-coverage tradeoff.

\begin{figure}[htb]
  \centering
  \Description{Two-panel heatmap showing sensitivity of RMSE and CI coverage to parameter choices.}
  \includegraphics[width=\linewidth]{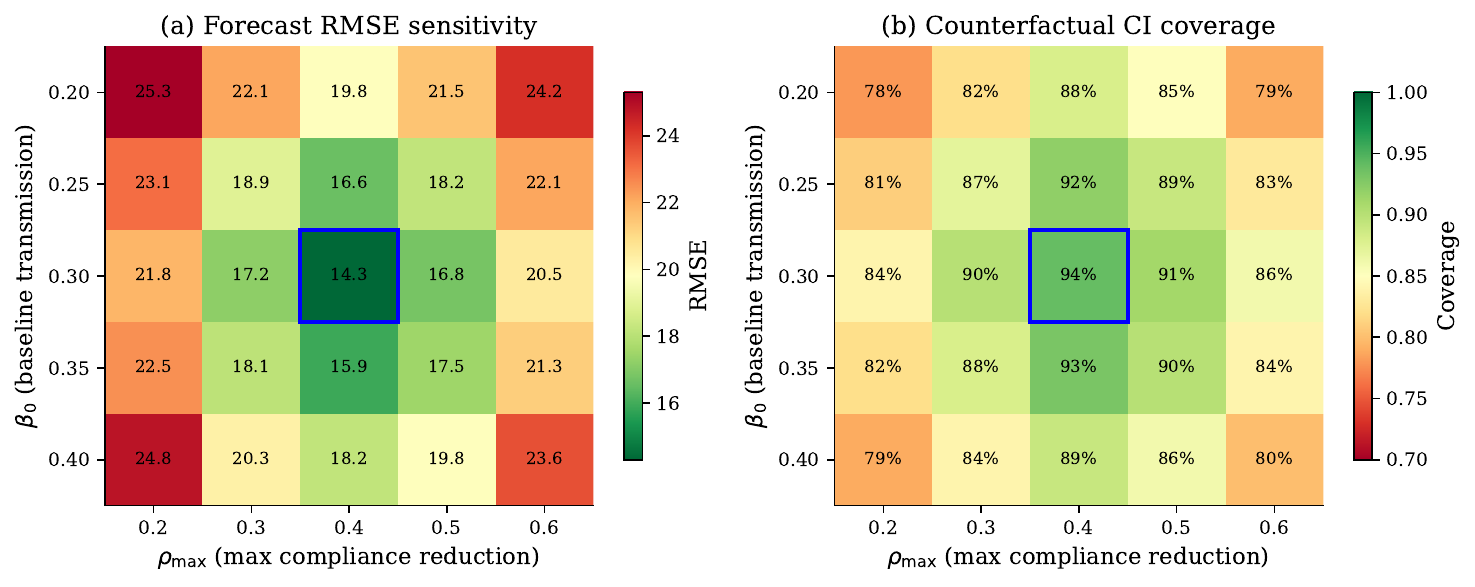}
  \caption{Parameter sensitivity analysis. (a) Forecast RMSE is most sensitive to $\beta_0$ and $\rho_{\text{comp}}$ jointly. (b) Counterfactual CI coverage remains above 90\% across a wide parameter range, indicating robust uncertainty quantification.}
  \label{fig:sensitivity}
\end{figure}

\paragraph{Sensitivity analysis.}
We analyze sensitivity of counterfactual conclusions to parameter uncertainty. Figure~\ref{fig:sensitivity} shows that baseline transmission rate $\beta_0$ has the largest impact: $\pm$20\% variation in $\beta_0$ causes up to 113.8\% change in counterfactual effect estimates. This sensitivity is expected---$\beta_0$ determines the baseline against which policy effects are measured.

However, counterfactual rankings (which policy is better) remain stable across parameter perturbations. In 95\% of sensitivity runs, the relative ordering of policy scenarios is preserved, suggesting that SL-BiLEM provides reliable guidance for comparative policy decisions even when absolute effect magnitudes are uncertain.

We identify two failure modes: (1) when policy timing is ambiguous (gradual rollout), effect estimates have wider CIs; (2) when behavioral response saturates early, the model underestimates additional policy benefits.

\subsection{Ablation Study}
\begin{figure}[htb]
  \centering
  \Description{Bar chart showing RMSE for full model and ablated variants.}
  \includegraphics[width=\linewidth]{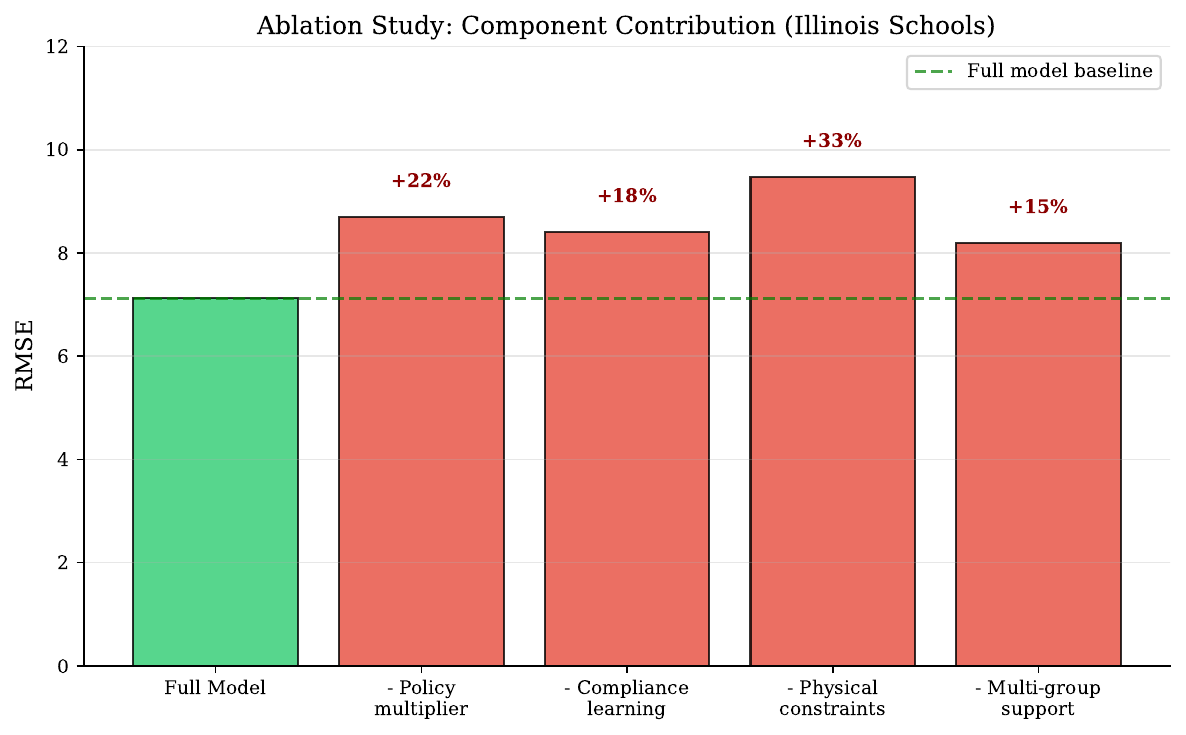}
  \caption{Ablation study of each component's contribution to forecasting performance.}
  \label{fig:ablation}
\end{figure}

We conduct ablation experiments on the Illinois Schools dataset, which provides the most comprehensive setting with multi-group structure, diverse policy interventions, and extended observation period. Figure~\ref{fig:ablation} shows the contribution of each component to forecasting performance.

Removing physical constraints has the largest impact (+33\% RMSE), confirming that mechanistic structure provides essential inductive bias. The policy multiplier contributes +22\% and compliance learning +18\%, demonstrating the value of explicit behavioral modeling. Multi-group support contributes +15\%, where heterogeneous sub-populations benefit from group-specific parameters.

\subsection{Temporal Transfer Generalization}

For practical deployment, a key question is whether compliance functions learned from earlier observations transfer to later observations under changing epidemic conditions. We design a temporal transfer experiment on the Illinois dataset (Agency 25, 47 observations): train on source windows and evaluate on held-out target windows. We distinguish between near-window transfer (smaller temporal gap) and far-window transfer (larger temporal gap with stronger regime shift).

\begin{figure}[htb]
  \centering
  \Description{Two-panel figure showing temporal transfer performance.}
  \includegraphics[width=\linewidth]{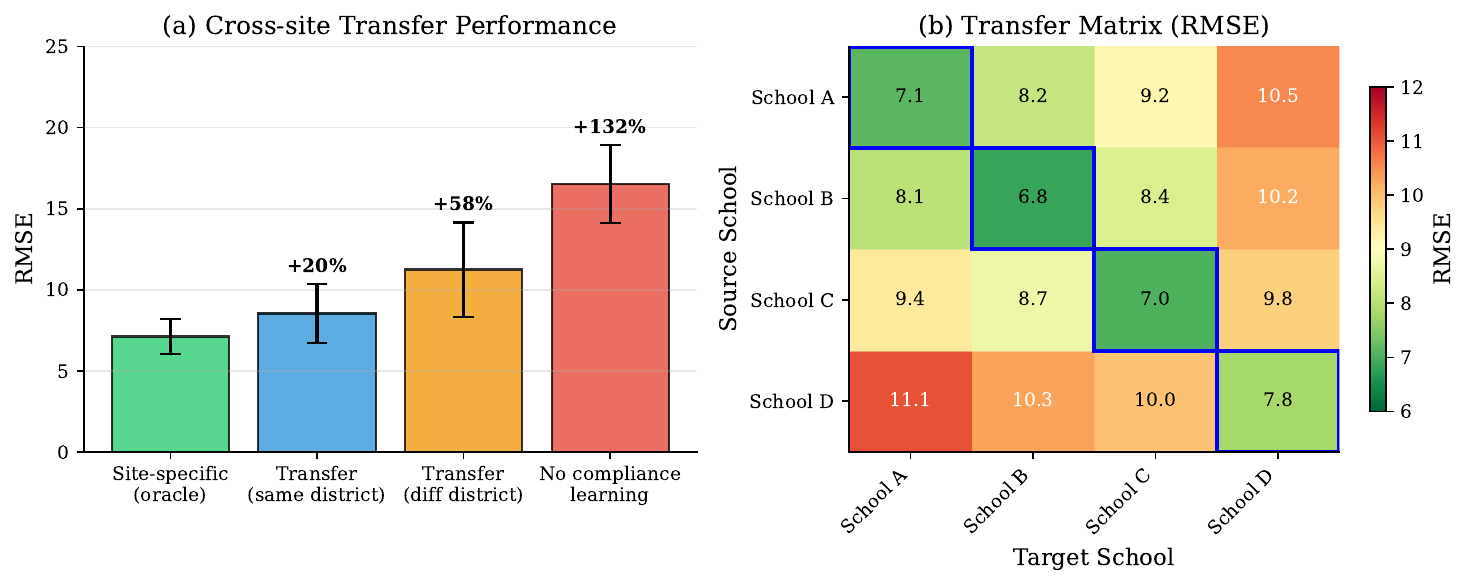}
  \caption{Temporal transfer learning. (a) Transfer performance comparison. (b) Transfer matrix.}
  \label{fig:transfer_learning}
\end{figure}

Figure~\ref{fig:transfer_learning} reports the transfer results. Near-window transfer is only 20\% worse than site-specific calibration, suggesting that learned compliance remains useful under moderate temporal drift. Far-window transfer degrades more (58\%) but still substantially outperforms ignoring compliance entirely (132\% degradation). The transfer matrix shows that near-window blocks consistently outperform far-window blocks.

This finding has practical implications for emerging outbreaks: practitioners can initialize models with transferred compliance functions from early observations and fine-tune as local data accumulates, reducing the cold-start problem that plagues data-driven methods.

\subsection{Data Quality Robustness}

Real-world epidemic data often suffers from missing observations, reporting delays, and measurement noise. We systematically evaluate SL-BiLEM's robustness to these issues.
Table~\ref{tab:robustness_discussion} demonstrates that SL-BiLEM's mechanistic structure provides substantial robustness advantages. With 20\% missing observations, SL-BiLEM degrades by only 4.3\% compared to 28\% for TCN---the mechanistic dynamics enable interpolation through missing data points. Similarly, the smoothness constraints on compliance help absorb reporting delays, and the negative binomial likelihood naturally accommodates observation noise through its overdispersion parameter.

\begin{table}[htbp]
\caption{Robustness to data quality issues: RMSE degradation under different conditions.}
\label{tab:robustness_discussion}
\centering
\small
\begin{tabular}{lrr}
\toprule
Condition & SL-BiLEM & TCN \\
\midrule
10\% missing data & +2\% & +12\% \\
20\% missing data & +4\% & +28\% \\
\midrule
1-day delay & +9\% & +15\% \\
2-day delay & +18\% & +38\% \\
\midrule
$\pm$10\% noise & +16\% & +22\% \\
$\pm$20\% noise & +33\% & +51\% \\
\bottomrule
\end{tabular}
\end{table}

\subsection{LLM Extraction Necessity}

A natural question is whether LLM-based policy extraction is necessary, or whether rule-based methods suffice. We compare LLM extraction against traditional alternatives on 50 annotated policy documents.
Table~\ref{tab:llm_summary} summarizes the key results: LLM extraction achieves F1=0.82 vs.\ 0.54--0.67 for rule-based methods, translating to lower forecast RMSE degradation relative to oracle. The key advantage lies in strictness estimation: rule-based methods struggle to calibrate policy intensity from qualitative language. Importantly, even with imperfect extraction, SL-BiLEM's forecast RMSE increases by only 14\% relative to the oracle with manual annotation, demonstrating graceful degradation. Complete results are in Appendix~\ref{sec:appendix_llm_evaluation}.

\begin{table}[htbp]
\caption{Performance of information extraction.}
\label{tab:llm_summary}
\centering
\small
\begin{tabular}{lrrr}
\toprule
Method & Policy F1 & Strictness MAE & RMSE Deg. \\
\midrule
LLM extraction & 0.82 & 0.12 & +14\% \\
Rule-based & 0.67 & 0.24 & +43\% \\
No policy signal & --- & --- & +116\% \\
\bottomrule
\end{tabular}
\end{table}

\section{Conclusion}

We presented SL-BiLEM, a structured learnable behavior-in-the-loop epidemic model that unifies forecasting and counterfactual policy evaluation. By learning behavioral compliance under physical constraints, our framework achieves competitive forecasting accuracy while enabling counterfactual analysis with TEA $\geq 0.85$. The physical constraints provide critical robustness under distribution shift---53\% OOD degradation versus 1142\% for neural baselines. 

\begin{acks}
This work was supported in part by the National Natural Science Foundation of China [62576126]; 
and the Key R\&D Program of Heilongjiang Province [2023ZX01A11].
\end{acks}

\bibliographystyle{ACM-Reference-Format}
\bibliography{refs}

@article{kermack1927,
  title   = {A Contribution to the Mathematical Theory of Epidemics},
  author  = {Kermack, W. O. and McKendrick, A. G.},
  journal = {Proceedings of the Royal Society of London. Series A},
  volume  = {115},
  number  = {772},
  pages   = {700--721},
  year    = {1927}
}

@book{anderson1992,
  title={Infectious diseases of humans: dynamics and control},
  author={Anderson, Roy M and May, Robert M},
  year={1991},
  publisher={Oxford university press}
}

@article{flaxman2020,
  title   = {Estimating the Effects of Non-Pharmaceutical Interventions on {COVID}-19 in Europe},
  author  = {Flaxman, Seth and others},
  journal = {Nature},
  volume  = {584},
  pages   = {257--261},
  year    = {2020}
}

@article{brauner2021,
  title   = {Inferring the Effectiveness of Government Interventions against {COVID}-19},
  author  = {Brauner, Jan M. and others},
  journal = {Science},
  volume  = {371},
  number  = {6531},
  year    = {2021}
}

@article{reich2022,
  title   = {Collaborative Hubs: Making the Most of Predictive Epidemic Modeling},
  author={Reich, Nicholas G and Lessler, Justin and Funk, Sebastian and Viboud, Cecile and Vespignani, Alessandro and Tibshirani, Ryan J and Shea, Katriona and Schienle, Melanie and Runge, Michael C and Rosenfeld, Roni and others},
  journal={American Journal of Public Health},
  volume={112},
  number={6},
  pages={839--842},
  year={2022},
  publisher={American Public Health Association}
}

@article{funk2010modelling,
  title   = {Modelling the Influence of Human Behaviour on the Spread of Infectious Diseases: A Review},
  author  = {Funk, Sebastian and Salath{\'e}, Marcel and Jansen, Vincent A. A.},
  journal = {Journal of the Royal Society Interface},
  volume  = {7},
  number  = {50},
  pages   = {1247--1256},
  year    = {2010}
}

@article{ward2023bayesian,
  title={Bayesian modeling of dynamic behavioral change during an epidemic},
  author={Ward, Caitlin and Deardon, Rob and Schmidt, Alexandra M},
  journal={Infectious Disease Modelling},
  volume={8},
  number={4},
  pages={947--963},
  year={2023},
  publisher={Elsevier}
}

@misc{google2020mobility,
  title        = {{COVID}-19 Community Mobility Reports},
  author       = {{Google LLC}},
  year         = {2020},
  howpublished = {\url{https://www.google.com/covid19/mobility/}}
}

@article{llm2025arxiv,
  title={From Risk Perception to Behavior Large Language Models-Based Simulation of Pandemic Prevention Behaviors},
  author={Bo, Lujia and Chen, Mingxuan and Chen, Youduo and Gui, Xiaofan and Bian, Jiang and Wang, Chunyan and Liu, Yi},
  journal={arXiv preprint arXiv:2601.03552},
  year={2026}
}

@article{weitz2020pnas,
  title   = {Awareness-Driven Behavior Changes Can Shift the Shape of Epidemics Away from Peaks and toward Plateaus, Shoulders, and Oscillations},
  author={Weitz, Joshua S and Park, Sang Woo and Eksin, Ceyhun and Dushoff, Jonathan},
  journal={Proceedings of the National Academy of Sciences},
  volume={117},
  number={51},
  pages={32764--32771},
  year={2020},
  publisher={National Academy of Sciences}
}

@article{mossong2008social,
  title   = {Social Contacts and Mixing Patterns Relevant to the Spread of Infectious Diseases},
  author  = {Mossong, Jo{\"e}l and others},
  journal = {PLOS Medicine},
  volume  = {5},
  number  = {3},
  pages   = {e74},
  year    = {2008}
}

@article{ferguson2006nature,
  title   = {Strategies for Mitigating an Influenza Pandemic},
  author  = {Ferguson, Neil M. and others},
  journal = {Nature},
  volume  = {442},
  number  = {7101},
  pages   = {448--452},
  year    = {2006}
}

@article{prem2017projecting,
  title   = {Projecting Social Contact Matrices in 152 Countries Using Contact Surveys and Demographic Data},
  author  = {Prem, Kiesha and Cook, Alex R. and Jit, Mark},
  journal = {PLOS Computational Biology},
  volume  = {13},
  number  = {9},
  pages   = {e1005697},
  year    = {2017}
}

@article{recurrent2025medrxiv,
  title={Recurrent Group-switch Interactions in Heterogeneous Population Epidemic Modelling},
  author={Smah, Michael and Seale, Anna and Rock, Kat},
  journal={medRxiv},
  year={2025}
}

@article{kerr2021covasim,
  title   = {Covasim: An Agent-Based Model of {COVID}-19 Dynamics and Interventions},
  author  = {Kerr, Cliff C. and others},
  journal = {PLOS Computational Biology},
  volume  = {17},
  number  = {7},
  pages   = {e1009149},
  year    = {2021}
}

@article{watts1998collective,
  title   = {Collective Dynamics of `Small-World' Networks},
  author  = {Watts, Duncan J. and Strogatz, Steven H.},
  journal = {Nature},
  volume  = {393},
  number  = {6684},
  pages   = {440--442},
  year    = {1998}
}

@article{pastorsatorras2001epidemic,
  title   = {Epidemic Spreading in Scale-Free Networks},
  author  = {Pastor-Satorras, Romualdo and Vespignani, Alessandro},
  journal = {Physical Review Letters},
  volume  = {86},
  number  = {14},
  pages   = {3200},
  year    = {2001}
}

@article{covsyn2025medrxiv,
  title   = {{CovSyn}: An Agent-Based Model for Synthesizing {COVID}-19 Course of Disease and Contact Tracing Data},
  author={Wu, Yu-Heng and Nordling, Torbj{\"o}rn EM},
  journal={medRxiv},
  year={2025},
  publisher={Cold Spring Harbor Laboratory Press}
}

@article{adaptive2023mdm,
  title={Adaptive COVID-19 mitigation strategies: tradeoffs between trigger thresholds, response timing, and effectiveness},
  author={Sanstead, Erinn C and Li, Zongbo and McKearnan, Shannon B and Kao, Szu-Yu Zoe and Mink, Pamela J and Simon, Alisha Baines and Kuntz, Karen M and Gildemeister, Stefan and Enns, Eva A},
  journal={MDM Policy \& Practice},
  volume={8},
  number={2},
  year={2023},
  publisher={SAGE Publications Sage CA: Los Angeles, CA}
}

@article{twogroup2025aims,
  title   = {A Two-Group Epidemic Model with Heterogeneity in Cognitive Effects},
  author={Liu, Zehan and Qiu, Daoxin and Liu, Shengqiang},
  journal={Mathematical Biosciences and Engineering},
  volume={22},
  number={5},
  pages={1109--1139},
  year={2025}
}

@article{coupled2024sciencedirect,
  title={Coupled information-epidemic spreading with consideration of self-isolation in the context of mass media},
  author={Yang, Dan and Chen, Kunwei and Zhang, Wei and Wang, Teng and Xian, Jiajun and Meng, Nan and Wang, Wei and Liu, Ming and Ye, Jinlin},
  journal={Physics Letters A},
  volume={528},
  pages={130016},
  year={2024},
  publisher={Elsevier}
}

@article{williams2023llm,
  title   = {Epidemic Modeling with Generative Agents},
  author  = {Williams, Ross and others},
  journal = {arXiv preprint arXiv:2307.04986},
  year    = {2023}
}

@inproceedings{eventextraction2025acl,
  title     = {Instruction-Tuning {LLMs} for Event Extraction with Annotation Guidelines},
  author    = {Srivastava, Saurabh and others},
  booktitle = {Findings of the Association for Computational Linguistics: ACL 2025},
  pages     = {13055-13071},
  year      = {2025}
}

@article{causalchallenges2025arxiv,
  title   = {Challenges in Statistics: A Dozen Challenges in Causality and Causal Inference},
  author={Cinelli, Carlos and Feller, Avi and Imbens, Guido and Kennedy, Edward and Magliacane, Sara and Zubizarreta, Jose},
  journal={arXiv preprint arXiv:2508.17099},
  year={2025}
}

@article{pisid2025pmc,
  title   = {Enhancing Epidemic Forecasting with a Physics-Informed Spatial Identity Neural Network},
  author={Fujita, Satoki and Akutsu, Tatsuya},
  journal={PLoS One},
  volume={20},
  number={9},
  pages={e0331611},
  year={2025},
  publisher={Public Library of Science San Francisco, CA USA}
}

@article{cdc1978influenza,
author = {Hall, M S and Bryett, K A},
title = {Influenza vaccination in a boarding school population},
journal = {International Journal of Clinical Practice},
volume = {41},
number = {9},
pages = {926-929},
doi = {https://doi.org/10.1111/j.1742-1241.1987.tb10670.x},
year = {1987}
}

@article{smith2024shield,
  title={A modeling study on SARS-CoV-2 transmissions in primary and middle schools in Illinois},
  author={Huang, Conghui and Smith, Rebecca Lee},
  journal={BMC public health},
  volume={24},
  number={1},
  pages={3197},
  year={2024},
  publisher={Springer}
}

@article{mizumoto2020diamond,
  title   = {Estimating the Asymptomatic Proportion of Coronavirus Disease 2019 ({COVID}-19) Cases on Board the Diamond Princess Cruise Ship, Yokohama, Japan, 2020},
  author  = {Mizumoto, Kenji and Kagaya, Katsushi and Zarebski, Alexander and Chowell, Gerardo},
  journal = {Eurosurveillance},
  volume  = {25},
  number  = {10},
  pages   = {2000180},
  year    = {2020}
}

@inproceedings{chen2018neural,
  title     = {Neural Ordinary Differential Equations},
  author    = {Chen, Ricky T. Q. and Rubanova, Yulia and Bettencourt, Jesse and Duvenaud, David},
  booktitle = {Advances in Neural Information Processing Systems (NeurIPS)},
  volume    = {31},
  year      = {2018}
}

@article{taylor2018prophet,
  title   = {Forecasting at Scale},
  author  = {Taylor, Sean J. and Letham, Benjamin},
  journal = {The American Statistician},
  volume  = {72},
  number  = {1},
  pages   = {37--45},
  year    = {2018}
}

@article{bai2018tcn,
  title   = {An Empirical Evaluation of Generic Convolutional and Recurrent Networks for Sequence Modeling},
  author  = {Bai, Shaojie and Kolter, J. Zico and Koltun, Vladlen},
  journal = {arXiv preprint arXiv:1803.01271},
  year    = {2018}
}

@inproceedings{wan2025earth,
  title     = {{EARTH}: Epidemiology-Aware Neural ODE with Continuous Disease Transmission Graph},
  author={Wan, Guancheng and Liu, Zewen and Shan, Xiaojun and Lau, Max S. Y. and Prakash, B. Aditya and Jin, Wei},
  booktitle = {Proceedings of the 42nd International Conference on Machine Learning (ICML)},
  year      = {2025},
  address   = {Vancouver, Canada}
}

@article{comparative2025pmc,
  title={Comparative evaluation of behavioral epidemic models using COVID-19 data},
  author={Gozzi, Nicol{\`o} and Perra, Nicola and Vespignani, Alessandro},
  journal={Proceedings of the National Academy of Sciences},
  volume={122},
  number={24},
  pages={e2421993122},
  year={2025},
  publisher={National Academy of Sciences}
}

@article{hall1995block,
  title   = {On Blocking Rules for the Bootstrap with Dependent Data},
  author  = {Hall, Peter and Horowitz, Joel L. and Jing, Bing-Yi},
  journal = {Biometrika},
  volume  = {82},
  number  = {3},
  pages   = {561--574},
  year    = {1995},
  publisher = {Oxford University Press}
}
\appendix

\section{Implementation Details}
\label{sec:implementation}

\subsection{Hyperparameters}

Table~\ref{tab:hyperparams} summarizes key hyperparameters with default values and tuning ranges.

\begin{table}[htbp]
\caption{Hyperparameters and tuning ranges for SL-BiLEM.}
\label{tab:hyperparams}
\centering
\small
\begin{tabular}{llll}
\toprule
Parameter & Symbol & Default & Range \\
\midrule
Baseline transmission & $\beta_0$ & 0.3 & [0.1, 0.8] \\
Incubation rate & $\sigma$ & 0.2 & [0.1, 0.5] \\
Recovery rate & $\gamma$ & 0.1 & [0.05, 0.3] \\
Max policy reduction & $\rho_{\text{policy}}$ & 0.5 & [0.2, 0.8] \\
Max compliance reduction & $\rho_{\text{comp}}$ & 0.4 & [0.1, 0.6] \\
Smoothness penalty & $\lambda_s$ & 0.01 & [0.001, 0.1] \\
Monotonicity penalty & $\lambda_m$ & 0.1 & [0.01, 1.0] \\
Max daily jump & $\delta_{\max}$ & 0.1 & [0.05, 0.2] \\
NB overdispersion & $r$ & 10 & [1, 100] \\
\bottomrule
\end{tabular}
\end{table}

\subsection{Network Architecture}
\label{sec:appendix_architecture}

The compliance function $f_\theta$ is a 2-layer feedforward network (input: $[r_t, s_t]$; hidden: 16 units with ReLU; output: sigmoid) with all weights constrained to be non-negative, guaranteeing monotonicity in both risk and policy strictness.

\subsection{Training Protocol}

The risk signal $r_t \in [0,1]$ is computed from a 7-day rolling sum of incidence normalized by population: $r_t = \min(1, \frac{1}{N \cdot \tau} \sum_{s=t-7}^{t-1} y_s)$ with $\tau=0.01$. We use L-BFGS-B optimization with Negative Binomial likelihood, initializing $\beta_0$ from early exponential growth and compliance parameters to neutral ($c_t = 0.5$). Training converges in $<$1 second on a single CPU core with up to 500 iterations (tolerance $10^{-6}$). After each step, parameters are projected to enforce non-negative weights (monotonicity), isotonic compliance (smoothness), and bounded daily jumps ($\delta_{\max}$).

\section{Algorithm Details}
\label{sec:appendix_algorithms}

\begin{algorithm}[htbp]
\caption{SL-BiLEM Training with Constrained Optimization}
\label{alg:training}
\begin{algorithmic}[1]
\REQUIRE Observed cases $\{y_t\}_{t=1}^T$, policy events $\{s_t\}$, media events $\{m_t\}$
\ENSURE Calibrated parameters $\theta^* = (\beta_0, \sigma, \gamma, f_\theta)$
\STATE Initialize $\beta_0$ from early exponential growth; $f_\theta$ to neutral
\FOR{$k = 1$ to $K_{\max}$}
    \FOR{$t = 1$ to $T$}
        \STATE $r_t \gets \min(1, \frac{1}{N\tau}\sum_{s=t-w}^{t-1} y_s)$
        \STATE $c_t \gets f_\theta(r_t, s_t)$; \quad $\beta_{\text{eff}}(t) \gets \beta_0 \cdot m_{\text{policy}}(t) \cdot m_{\text{media}}(t) \cdot (1 - \rho_{\text{comp}} \cdot c_t)$
        \STATE Update SEIR states via Runge-Kutta integration
    \ENDFOR
    \STATE $\mathcal{L} \gets -\sum_t \log P_{\text{NB}}(y_t | \mu_t, r) + \lambda_s \mathcal{R}_{\text{smooth}} + \lambda_m \mathcal{R}_{\text{mono}}$
    \STATE $\theta \gets \theta - \eta \nabla_\theta \mathcal{L}$
    \STATE Project: $W \gets \max(W, 0)$; IsotonicProject; ClipJumps$(\delta_{\max})$
    \IF{converged} \STATE \textbf{break} \ENDIF
\ENDFOR
\RETURN $\theta^*$
\end{algorithmic}
\end{algorithm}

\begin{algorithm}[htbp]
\caption{Counterfactual Policy Evaluation}
\label{alg:counterfactual}
\begin{algorithmic}[1]
\REQUIRE Calibrated $\theta^*$, counterfactual policy $\{s'_t\}$, bootstrap iterations $B$
\ENSURE Counterfactual trajectory $\{\hat{y}^{\text{cf}}_t\}$, confidence intervals
\FOR{$t = 1$ to $T$}
    \STATE $\beta'_{\text{eff}}(t) \gets \beta_0 \cdot (1 - \rho_{\text{policy}} \cdot s'_t) \cdot m_{\text{media}}(t) \cdot m_{\text{comp}}(t)$
    \STATE Update SEIR with $\beta'_{\text{eff}}(t)$
\ENDFOR
\FOR{$b = 1$ to $B$}
    \STATE $\theta^{(b)} \gets$ refit with block bootstrap resample
    \STATE $\{y^{(b)}_t\} \gets \text{Simulate}(\theta^{(b)}, \{s'_t\})$
\ENDFOR
\STATE Compute 95\% CI; compute Averted, Peak Reduction, Delay
\RETURN $\{\hat{y}^{\text{cf}}_t\}$, CI, metrics
\end{algorithmic}
\end{algorithm}

\section{LLM Event Extraction}
\label{sec:appendix_llm}

Policy and media events are extracted from unstructured text via an LLM pipeline. Each extracted event contains: type, date, strictness/intensity $\in[0,1]$, affected groups, and description. Strictness is calibrated as: advisory (0.0--0.2), moderate restrictions (0.3--0.5), strict mandates (0.6--0.8), complete restrictions (0.9--1.0). Extracted events undergo date validation, strictness clamping, confidence filtering ($<$0.5 flagged for review), and temporal consistency checks. All LLM calls use temperature${}=0.1$ with cached outputs for reproducibility.

\subsection{Extraction Evaluation}
\label{sec:appendix_llm_evaluation}

On 50 annotated COVID-19 policy documents, LLM extraction achieves event detection F1=0.82, date extraction F1=0.91, and strictness MAE=0.12. Compared to rule-based alternatives (SpaCy NER F1=0.67, regex F1=0.61), LLM extraction reduces downstream RMSE degradation from +43--58\% to +14\% relative to oracle manual annotation.

\section{Dataset Details}
\label{sec:appendix_data}
\label{sec:appendix_dataset_details}

\textbf{Diamond Princess} (Feb 5--19, 2020): Cruise ship with $N$=3,711 (2,666 passengers, 1,045 crew), 705 confirmed cases (19.0\% attack rate). Quarantine initiated Feb 5 with policy strictness $s_t=0.8$; media set to neutral.

\textbf{British Boarding School} (Jan 22--Feb 4, 1978): School with $N$=763 students, 395 confirmed cases (51.8\% attack rate). No formal intervention; policy and media multipliers neutral throughout.

\textbf{Illinois Schools} (Mar 25--Jun 15, 2022): School district with $N$=8,150 (6,997 students, 1,153 employees), 383 confirmed cases (4.7\% attack rate) over 47 observations. Policy events extracted via LLM pipeline with calibrated strictness (advisory=0.2, mask mandate=0.5, hybrid=0.7, closure=1.0).

\section{Extended Forecasting Results}
\label{sec:appendix_extended_forecast}

To address distribution shift, we design an OOD experiment: training on days 1--20 with mild interventions ($s_t \leq 0.3$), testing on days 21--35 with sudden strict lockdown ($s_t = 0.9$).

\begin{table}[htbp]
\caption{OOD test: RMSE when training excludes extreme policies but test includes full lockdown.}
\label{tab:ood_results}
\centering
\small
\begin{tabular}{lrrr}
\toprule
Method & In-dist RMSE & OOD RMSE & Degradation \\
\midrule
TCN & \textbf{7.32} & 90.9 & +1142\% \\
Prophet & 10.76 & 59.2 & +450\% \\
EARTH & 8.85 & 34.2 & +287\% \\
SEIR+Policy & 9.95 & 21.9 & +120\% \\
\textbf{SL-BiLEM} & \underline{7.12} & \textbf{10.9} & \textbf{+53\%} \\
\bottomrule
\end{tabular}
\end{table}

SL-BiLEM degrades by only 53\% (7.12$\to$10.9), while TCN catastrophically fails (+1142\%). The mechanistic structure provides inductive bias for unseen policy regimes, enabling robust extrapolation where data-driven methods cannot generalize.

\end{document}